\documentclass[runningheads]{llncs}

\usepackage{hyperref}
\usepackage[hyphenbreaks]{breakurl}
\usepackage{amssymb}
\usepackage{amsmath}
\usepackage{graphicx}
\usepackage{booktabs}
\usepackage{makecell}
\usepackage{setspace}
\usepackage{subfigure} 
\usepackage{array} 
\usepackage{multirow}
\usepackage{bbding}
\usepackage[table,xcdraw]{xcolor}
\usepackage{amssymb}
\usepackage{pifont}
\usepackage{url}

\begin{document}
\title{ 
A Comprehensive Study on Knowledge Graph Embedding over Relational Patterns Based on Rule Learning}
\titlerunning{Comprehensive Study over Relational Patterns}

%

\author{
Long Jin\inst{1} 
\and
Zhen Yao\inst{1} 
\and
Mingyang Chen\inst{2} 
\and 
Huajun Chen\inst{2, 3} 
\and 
Wen Zhang\inst{1}\thanks{Corresponding author.} 
}
\authorrunning{L. Jin et al.}

\institute{School of Software Technology, Zhejiang University  \and 
College of Computer Science and Technology, Zhejiang University \and Donghai laboratory
\\
\email{\{longjin,yz0204,mingyangchen,huajunsir,zhang.wen\}@zju.edu.cn}}

\maketitle              
\begin{abstract}

Knowledge Graph Embedding (KGE) has proven to be an effective approach to solving the Knowledge Graph Completion (KGC) task. Relational patterns which refer to relations with specific semantics exhibiting graph patterns are an important factor in the performance of KGE models. Though KGE models' capabilities are analyzed over different relational patterns in theory and a rough connection between better relational patterns modeling and better performance of KGC has been built, a comprehensive quantitative analysis on KGE models over relational patterns remains absent so it is uncertain how the theoretical support of KGE to a relational pattern contributes to the performance of triples associated to such a relational pattern. To address this challenge, we evaluate the performance of 7 KGE models over 4 common relational patterns on 2 benchmarks, then conduct an analysis in theory, entity frequency, and part-to-whole three aspects and get some counterintuitive conclusions. Finally, we introduce a training-free method Score-based Patterns Adaptation (SPA) to enhance KGE models' performance over various relational patterns. This approach is simple yet effective and can be applied to KGE models without additional training. Our experimental results demonstrate that our method generally enhances performance over specific relational patterns. Our source code is available from GitHub at \url{https://github.com/zjukg/Comprehensive-Study-over-Relational-Patterns}.


\end{abstract}
\section{Introduction}
Knowledge Graphs (KGs) are used to organize triples and represent various types of information about the real world. A typical triple consists of a head entity, a relation, and a tail entity, expressed in the format $(h, r, t)$. Several well-known KG projects, including FreeBase~\cite{bollacker2008freebase}, WordNet~\cite{miller1995wordnet}, YAGO~\cite{suchanek2007yago}, and DBpedia~\cite{lehmann2015dbpedia}, have gained attention for their successful use in natural language processing~\cite{yang2019leveraging}, question answering~\cite{bordes2014open}, recommendation systems~\cite{wang2019explainable}, and other downstream tasks.

Despite the vast number of triples in large-scale KGs, they suffer from the problem of incompleteness. To address this problem, the Knowledge Graph Completion (KGC) task, such as link prediction, aims to predict missing triples based on known triples. Knowledge Graph Embedding (KGE)~\cite{wang2017knowledge} has proven to be an effective approach to solving the KGC task by capturing semantic representations of entities and relations in a low-dimensional vector space.


Most KGE methods utilizing triples as learning resources derive the semantics of entities and relations from graph structures.
Relations with specific semantics typically exhibit corresponding graph patterns,
which we call relational patterns in this paper, such as symmetry/antisymmetry, inversion, and composition~\cite{sun2019rotate}. 
The performance of KGE models is widely regarded as being closely tied to their capacity for capturing relational patterns within the KG~\cite{zhang2019iteratively,sun2019rotate,qu2019probabilistic,li2021there,cao2021dual}.
Previous studies have endeavored to explore whether KGE models truly learned the relational patterns among triples. 
Some works~\cite{yang2014embedding,zhang2019iteratively} use the learned relation embeddings to mine rules corresponding to different relational patterns and prove that the mined rules are of high quality, showing KGE models successfully learned the relational patterns among triples.
Some works~\cite{sun2019rotate,zhang2020learning} utilize distribution histograms to reveal that embeddings of relations associated with relational patterns tend to converge towards specific positions within the vector space. 
Although previous studies have theoretically analyzed KGE models' capabilities in addressing various relational patterns and established a rough connection between better relational patterns modeling and better performance of KGC, a comprehensive quantitative analysis of KGE models to relational patterns remains absent. In the absence of such research, it is uncertain how the theoretical support of KGE for a specific relational pattern contributes to the prediction results of triples associated with that pattern. Consequently, quantifying KGE models' performance over particular relational patterns poses a significant challenge.



In this paper, 
we propose a methodology to classify triples into relational patterns based on rules mined from training data, then the capacity of KGE models in reasoning over different patterns can be quantified with the performance of triples belonging to specific patterns.
We conduct numerous experiments, in theory, entity frequency, and part-to-whole aspects, to assess KGE models' performance over relational patterns, leading to the following conclusions:
\textbf{1)} Theoretical support for a relational pattern in a KGE model does not guarantee superior performance compared to another KGE model lacking such support.
\textbf{2)} The influence of entity frequency on the performance of different relational patterns varies. Performance for symmetric patterns diminishes as entity frequency increases, while for other patterns, performance improves with increasing frequency.
\textbf{3)} If one KGE model significantly outperforms another, the superior model will exhibit better performance overall relational patterns. Conversely, when two KGE models exhibit similar overall performance, their performance over relational patterns may diverge considerably. Lastly, we introduce a training-free method, Score-based Patterns Adaptation (SPA), to enhance KGE models' performance over various relational patterns. 
Our experimental results demonstrate that SPA generally improves performance over specific relational patterns.

The contributions of this paper are summarized as follows:
\begin{enumerate}
    \item  To our best knowledge, we are the first to conduct a comprehensive quantitative analysis over relational patterns;
    \item  We evaluate the performance of 7 KGE models over 4 common relational patterns on 2 benchmarks, then provide an analysis in theory, entity frequency, and part-to-whole three aspects, and get some counterintuitive conclusions;
    \item  We introduce a training-free method, Score-based Patterns Adaptation (SPA), designed to enhance KGE models' performance over various relational patterns. This approach is simple yet effective and can be applied to KGE models without additional training.
\end{enumerate}

This article is structured as follows: Section~\ref{section2} introduces the related work and Section~\ref{section3} introduces preliminaries and background. Sections~\ref{section4} and ~\ref{section5} are the main part of the paper, presenting our methodology, the comprehensive quantitative analysis of patterns, and SPA results. Section~\ref{section6} presents conclusions and an outlook for future work.

\section{Related Work}\label{section2}

In this section, we concentrate on the related work of this paper, which encompasses the process of KGE, the definition of various relational patterns, and rule mining for relational patterns.

\paragraph{Knowledge Graph Embedding.}

Knowledge Graph Embedding (KGE) models strive to capture the semantic meanings of entities and relations by mapping them to continuous vectors, allowing for effective information retrieval and knowledge discovery. The process of KGE generally initializes the entity and relation embeddings and subsequently updates them with the score function and loss function. \textbf{Score function} measures the plausibility of a triple $(h, r, t)$ with embeddings. These functions can be classified into two categories: translational distance based and semantic matching based models~\cite{wang2017knowledge,ji2021survey}. The translational distance based model primarily includes TransE and its extensions (such as TransH~\cite{wang2014knowledge}, TransR~\cite{lin2015learning}, TransD~\cite{ji2015knowledge}), RotatE~\cite{sun2019rotate}, and others. Semantic matching based models mainly comprise RESCAL~\cite{nickel2011three}, DistMult~\cite{yang2014embedding}, ComplEx~\cite{trouillon2016complex} and more. Table~\ref{tab:score function} presents some common score functions of KGE models. \textbf{Negative sampling} is the process of generating negative samples as most KGs predominantly contain positive triples.
Negative triples are produced by corrupting a positive triple $(h, r, t)$ through the replacement of either $h$ or $t$. 
Established methods encompass uniform negative sampling~\cite{bordes2013translating}, Bernoulli negative sampling~\cite{wang2014knowledge}, and so on.
\textbf{Loss functions} strive to minimize the scores of negative triples while maximizing those of positive triples. 
Principal loss function methods mainly include pointwise logistic loss~\cite{trouillon2016complex}, pairwise hinge loss~\cite{bordes2013translating}, 
softplus loss~\cite{mohamed2019loss}, self-adversarial negative sampling loss~\cite{sun2019rotate} and others.


\paragraph{Relational Patterns.} 
Relational patterns serve as a crucial metric for evaluating the performance of KGE models. Our comprehension of relational patterns becomes deepening over time.
Wang et al.~\cite{wang2014knowledge} highlight several mapping properties of relations that ought to be considered when embedding a knowledge graph, including \textbf{reflexive}, \textbf{one-to-many}, \textbf{many-to-one}, and \textbf{many-to-many}. 
Xie et al.~\cite{xie2016representation} introduce a method called Type-embodied Knowledge Representation Learning (TKRL) that leverages \textbf{hierarchical entity types} to enhance the representation learning of knowledge graphs. 
Minervini et al.~\cite{minervini2017regularizing} incorporate \textbf{equivalence and inversion axioms} to improve the training of neural embeddings for knowledge graphs. 
The purpose of these axioms is to improve the accuracy and generalization abilities of neural embeddings by utilizing external background knowledge.
Sun et al.~\cite{sun2019rotate} discuss \textbf{relational patterns} such as \textbf{symmetry}/\textbf{antisymmetry}, \textbf{inversion}, and \textbf{composition}.
Then RotatE is proposed and demonstrates higher performance on various benchmarks.
Qu et al.~\cite{qu2019probabilistic} examine the \textbf{subrelational pattern} in the context of exploring the impact of different rule patterns on knowledge graph reasoning. 
Cao et al.~\cite{cao2021dual} suggest that \textbf{multiple relations} and propose DualE to model multiple relations using a combination of translation and rotation with greater performance. Some common relational patterns with their conditions are listed in Table~\ref{relational patterns}.

\paragraph{Rule Mining.}
Rule mining can be employed to uncover non-obvious structures in data with logical rules~\cite{zhang2022knowledge}. The logical rules serve as a flexible declarative language for conveying high-level cognition~\cite{hu2016harnessing,xu2022ruleformer}, which can enhance the accuracy of reasoning or contribute to the generation of new triples~\cite{cheng2021uniker,zhang2021explaining}.
Various rule mining methods have been developed to efficiently extract rules from large-scale knowledge graphs.
The WARMR~\cite{dehaspe1999discovery} and ALEPH~\cite{srinivasan2001aleph} discover association rules over a limited set of queries.
AMIE~\cite{galarraga2013amie} and Ontological Path-finding~\cite{chen2016ontological} mine rule based on an exhaustive top-down search with pruning strategies. AMIE+~\cite{galarraga2015fast} improves the precision of the forecasts by using joint reasoning and type information. RARL~\cite{pirro2020relatedness} uses relatedness between predicates to improve search efficiency. 
These mined logical rules can correspond to the majority of relational patterns we proposed (the relationships between them will be detailed in Section~\ref{section3}). 
In this paper, we employ the latest version of AMIE called AMIE3~\cite{lajus2020fast} to achieve length control and a trade-off between efficiency and quality.

\section{Preliminaries and Background}\label{section3}
\paragraph{Knowledge Graph.} 
With a set of entities $\mathcal{E}$ and a set of relations $\mathcal{R}$, Knowledge Graph $\mathcal{G}$ can be represented as a set of triplets $\mathcal{G} = \{ (h,r,t)\}$ in which $h \in \mathcal{E}$ and $t \in \mathcal{E}$ represent the head and tail entity respectively, $r \in \mathcal{R}$ represents the relationship between $h$ and $t$. A triple $(h,r,t)$ can also be represented as $r(h,t)$. Most KGs are far from complete. KGC
comes into play as a powerful application to infer missing links. For example, predicting the missing head or tail entities given $(h,r)$ or $(r,t)$ pairs.


\paragraph{Relational Patterns.}
We gave definitions of six key patterns~\cite{sun2019rotate,qu2019probabilistic} that could be written in regular rule form concerned in Table~\ref{relational patterns}, including symmetric, antisymmetric, inverse, equivalent, subrelation, and compositional patterns.

\begin{table}[]
\centering
\caption{Conditions for relational patterns and its rule formulation. $EN(r)$ refers to the set of entity pairs with $(head\ entity, tail\ entity)$ that satisfy $r(head\ entity, tail\ entity)$. $\emptyset$ refers to the empty set. The $r$ in each row is represented as the relation belonging to the corresponding pattern. Note that the $n$ in parentheses after the compositional indicates the number of relations in the hypothesis.}
\label{relational patterns}
\resizebox{.95\columnwidth}{!}{
\resizebox{\columnwidth}{!}{
\renewcommand\arraystretch{1.4} 
\begin{tabular}{@{}cllccll@{}}
\toprule
\multicolumn{3}{c}{\textbf{Relational   Patterns}} &
  \textbf{Condition} &
  \multicolumn{3}{c}{\textbf{Rule Form}} \\ \midrule
\multicolumn{3}{c}{\textbf{symmetric}} &
  $EN(r) = \{(t,h)|(h,t)\in EN(r) \}$ &
  \multicolumn{3}{c}{$ r(H,T) \leftrightarrow r(T,H) $} \\
\multicolumn{3}{c}{\textbf{antisymmetric}} &
  $ EN(r) \cap \{(t,h)|(h,t)\in EN(r) \} = \emptyset $ &
  \multicolumn{3}{c}{$ r(H,T) \nleftrightarrow r(T,H)$} \\
\multicolumn{3}{c}{\textbf{inverse}} &
  $EN(r') \subseteq \{(t,h)|(h,t)\in EN(r) \}$ &
  \multicolumn{3}{c}{$ r(H,T) \leftarrow r'(T,H) $} \\
\multicolumn{3}{c}{\textbf{equivalent}} &
  $EN(r') = EN(r)$ &
  \multicolumn{3}{c}{$r(H,T)\leftrightarrow r'(H,T)$} \\
\multicolumn{3}{c}{\textbf{subrelation}} &
  $EN(r') \subseteq EN(r)$ &
  \multicolumn{3}{c}{$ r(H,T) \leftarrow r'(H,T)$} \\
\multicolumn{3}{c}{\textbf{compositional(n)}} &
  \makecell[c]{$\{(h,t)|(h,a_1) \in EN(r_1)…$ \\ $(a_n,t) \in EN(r_n)\} \subseteq EN(r)$} &
  \multicolumn{3}{c}{\makecell[c]{$ r(H,T)$ \\ $\leftarrow r_1(H,X_1),... r_n(X_n,T)  $}} \\ \bottomrule
\end{tabular}}}
\end{table}

\paragraph{Closed-Path Rules}

The rule has the property of a closed path~\cite{yang2014embedding} if and only if the sequence in the hypotheses creates a path from the subject argument to the object argument of the conclusion predicate without any cycles or repeated nodes. Closed-path rules are a type of rule that is used to capture complex relationships between entities in KGs. A closed-path rule $\tau$ is of the form:
\begin{equation}
 r(H,T) \leftarrow r_1(H, X_1) \wedge r_2(X_1, X_2) \wedge ... \wedge r_n(X_{n-1}, T) 
\end{equation}
where $H,T,X_i$ are variables. We usually represent hypotheses $r_1(H, X_1) \wedge r_2(X_1, $\par \noindent $X_2) \wedge ... \wedge r_n(X_{n-1}, T)$ as the body and denote the conclusion $r(H,T)$ as the head of the rule. 
Rule quality could be evaluated through statistical metrics such as standard confidence, partial-close-world assumption (PCA) Confidence, and Head Coverage~\cite{galarraga2013amie}.


\begin{figure}[t]
\includegraphics[width=\columnwidth]{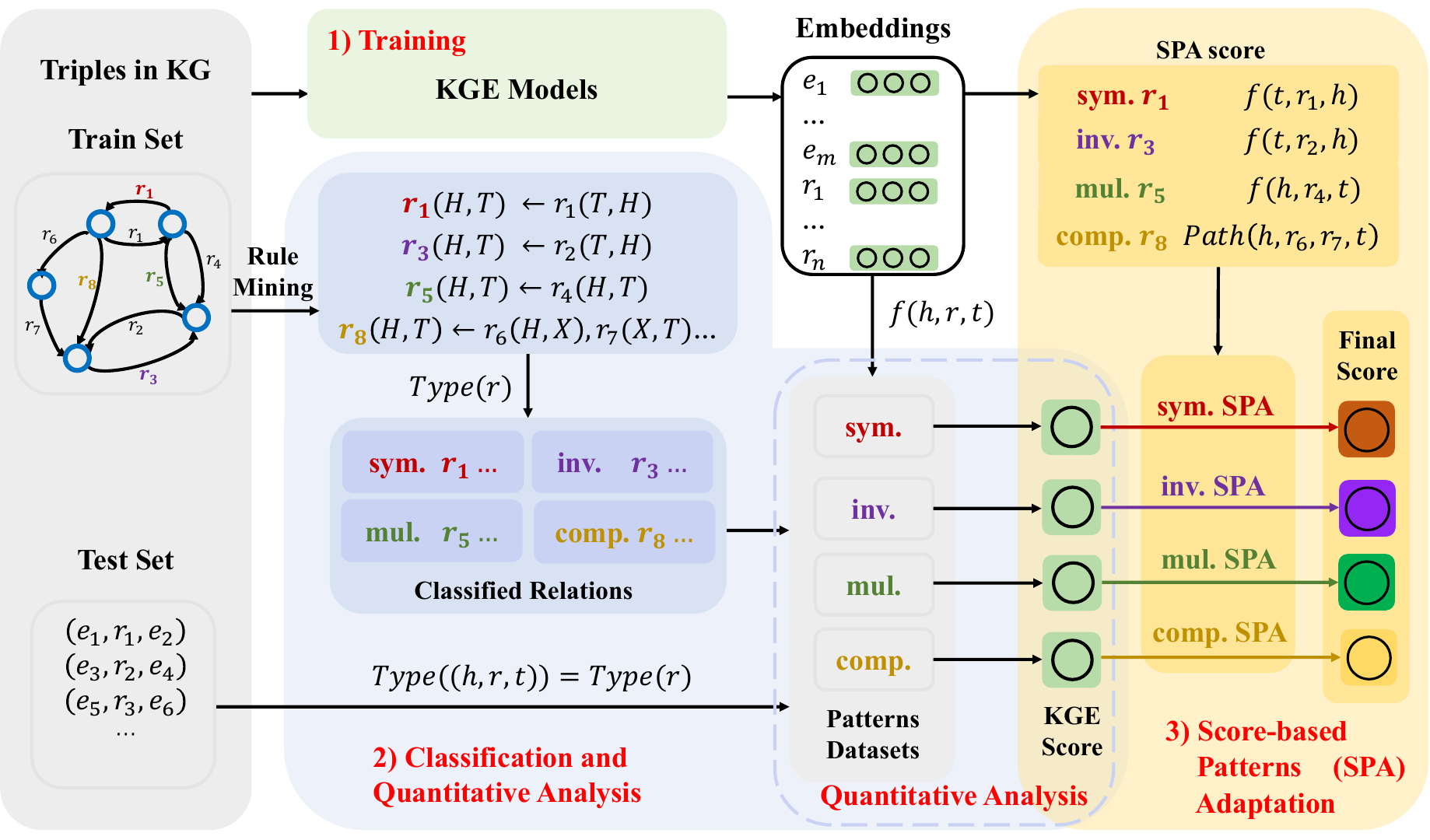}
\caption{The overall architecture of our methodology. Sym, inv, mul, comp2, and comp3 are abbreviations for symmetric, inverse, multiple, compositional2, and compositional3 respectively.
}

\label{overview}
\end{figure}

\section{Methodology}\label{section4}
The overall architecture of our methodology is shown in Fig.~\ref{overview}. Our work pipeline can be divided into the following three steps. \textbf{1)} In the training stage, train the KGE model on the training set to get entities' and relations' embeddings. \textbf{2)} In the classification and analysis stage, classify triple $(h,r,t)$ into patterns datasets based on the classification of $r$. Then the performance of different KGE models over different patterns is analyzed comprehensively and quantitatively through experiments on specific pattern datasets. 
\textbf{3)} Compute the final score by combining the SPA score and KGE score without requiring additional training.

The training of KGE models in step 1 follows the standard process, thus we omit the training details of KGE models and refer readers to the original papers of these methods. Next, we will describe Steps 2, and 3 of the architecture of our methodology in detail.

\subsection{Classification of Triples}

In this subsection, we focus on classifying triples into different relational pattern datasets to realize the comprehensive quantitative capabilities analysis of KGE models over various relational patterns. We first give a more detailed explanation of the classification method. Then we make an analysis of the relational pattern from the perspective of rule form and KG's property. 

\paragraph{Classification Method}
The rule form of relational patterns often takes the form of Horn rules with closed paths (as seen in Table~\ref{relational patterns}), which are similar to the rules mined by rule mining in KGs. Therefore, we decide to utilize rule mining to classify relations $Type(r)$ based on the rule form of distinct patterns. For example, if we get a rule $\tau: r(H, T) \leftarrow r(T, H)$ which is similar to the rule form of symmetric pattern $r(H, T) \leftrightarrow r(T, H)$, we will get $Type(r)=\{symmetric\}$ (note that one relation may belong to more than one pattern).
Classification of the triple $Type((h,r,t))$ is based on the classification of its relation $r$, in which $Type((h,r,t)) = Type(r)$.
The performance of the KGE model over 
a relational pattern $P$
can be quantified according to the 
performance of the set of triples belonging to this relational pattern that $\{ (h,r,t)| P \in Type((h,r,t))  \}$.


\paragraph{Relational Patterns Analysis}
For the relational patterns mentioned in Table~\ref{relational patterns}, we conduct a more detailed analysis of six relational patterns: 
\begin{enumerate}

\item  For the antisymmetric pattern, we consider it as the negative counterpart of symmetric patterns. Since the triples in knowledge graphs are all positive, the correctness of non-existing triples is unknown. Therefore, we only consider positive patterns in the experiments. 

\item Subrelation and equivalent patterns share a similar rule form. The test set is classified based on the rules obtained from the training set, with the test set used as the head of the rule to reversely get the body. However, since the training and test sets are randomly divided, the prior and posterior between them are weak. Thus, we merge these two patterns into \texttt{multiple} patterns, following the expression by Cao et al.~\cite{cao2021dual}. 

\item In the compositional pattern, if the length of the rule is $1$, it can become the previous pattern form. Some works~\cite{suresh2020hybrid,cheng2021uniker} incorporated language biases, such as restricting the length of rules up to 3 to deal with the vast search space and self-loops. We also set the length of rules to 2 and 3 in subsequent experiments.

\end{enumerate}

\subsection{Score-based Patterns Adaptation (SPA)}
To explicitly enhance the capability of KGE models over relational patterns, we aim to propose a simple approach to specifically promote KGE models on different patterns. Therefore, we introduce a training-free method, \textbf{Score-based Patterns Adaptation (SPA)}, which combines information from specific relational patterns and KGE score to modify the score function of models. 
The fundamental premise of SPA is that if the head triple is true, the triples within the body are also likely to be true with a high probability as the rule's head can be inferred from the body using established rules. 
Consequently, we propose to utilize the result derived from the rule's body as SPA score to enhance the inference over specific patterns. 
We also consider the rule confidence of rule mining as the confidence of relational patterns to measure the credibility of the modified score. The details are as follows:

\begin{table}[]

\centering
\caption{Correspondence between SPA score and rules in relational patterns. \texttt{Multiple} is the combination of subrelation and equivalent patterns mentioned in Table~\ref{relational patterns}.}
\label{SPAscore}
\resizebox{.85\columnwidth}{!}{
\renewcommand\arraystretch{1.25} 
\begin{tabular}{@{}cllcc@{}}
\toprule
\multicolumn{3}{c}{Relational Patterns} & Rule $\tau$   & SPA score $S_{p}(h,r,t)$ \\ \midrule
\multicolumn{3}{c}{\texttt{Symmetric}} & $ r(H,T) \leftarrow r(T,H) $    & $f(t,r,h)$  \\
\multicolumn{3}{c}{\texttt{Inverse}}   & $ r(H,T) \leftarrow r'(T,H)$  & $f(t,r',h)$ \\
\multicolumn{3}{c}{\texttt{Multiple}}  & $ r(H,T) \leftarrow r'(H,T)$  & $f(h,r',t)$ \\
\multicolumn{3}{c}{\texttt{Compositional}}  & $ r(H,T) \leftarrow r_1(H,X_1),... r_n(X_n,T) $ & $Path(h, r_1...r_n, t)$                \\ \bottomrule
\end{tabular}}

\end{table}

Table~\ref{SPAscore} shows the SPA score with rule $\tau$ on different relational patterns. Compared to Table 1, 
Table 2 is refined based on the relational patterns analysis in Table 1.
For the \texttt{symmetric} pattern, if we have the rule $\tau: r(H,T) \leftarrow r(T,H)$, we consider this relation $r$ in
the rule to be symmetric and can contribute to reasoning by calculating \texttt{symmetric} SPA score $f(t, r, h)$. And for the \texttt{inverse} pattern, we consider the relation $r$ to be inverse with relation $r'$ on rule $\tau: r(H,T) \leftarrow r'(T,H)$, and the semantic information in the body is $f(t, r', h)$ when computing the credibility of $(h,r,t)$. Similarly to \texttt{symmetric} and \texttt{inverse} patterns, we can get SPA score $f(h, r', t)$ for the \texttt{multiple} and $Path(h, r_1...r_n, t)$ for the \texttt{compositional}. 

Nevertheless, the compositional rules mined by the majority of existing open-source techniques are not readily applicable, largely due to their not chain-like structure~\cite{niu2020rule}.
Take the rule $ \tau: r(H, T) \leftarrow r_1(X, H), r_2(X, T) $ as an example, we need first convert the body part $r_1(X, H)$ into ${r_1}^{-1}(H, X)$, where ${r_1}^{-1}$ denotes the inverse relation of $r_1$, then we can obtain a compositional chain rule $r(H, T) \leftarrow {r_1}^{-1}(H, X), r_2(X, T) $, and see Appendix~\ref{sec:Compositional Pattern} for more details on the implementation of \texttt{Compositional} SPA.


\begin{table}[]
  \centering
  \caption{The details of KGE models, where $\| \cdot \|_1$ and $\| \cdot \|_2$ denote the absolute-value norm and Euclidean norm respectively. The expression of \textbf{h},\textbf{r},\textbf{t} in bold is denoted as their embedding vectors.}
  \resizebox{\columnwidth}{!}{
  \renewcommand\arraystretch{1.25} 
  \setlength{\tabcolsep}{2mm}{ 
  \label{tab:score function}
  \begin{tabular}{cccccc}
    \toprule
    Model & $f_{kge}(h, r, t)$ & $P a t h(h, r_1...r_n, t)$ \\
    \midrule
    TransE & $-\|\mathbf{h}+\mathbf{r}-\mathbf{t}\|_1$ & $-\|\mathbf{h}+\mathbf{r_1}...+\mathbf{r_n}-\mathbf{t}\|_1$ \\
    RotatE  & $-\|\mathbf{h} \circ \mathbf{r}-\mathbf{t}\|_2$ & $-\|\mathbf{h} \circ \mathbf{r_1} ... \circ \mathbf{r_n} -\mathbf{t}\|_2$ \\
    HAKE & \makecell[c]{ $ -\left\|\mathbf{h}_m \circ \mathbf{r}_m-\mathbf{t}_m\right\|_2- $ \\ $ \lambda\left\|\sin \left(\left(\mathbf{h}_p+\mathbf{r}_p-\mathbf{t}_p\right) / 2\right)\right\|_1 $ }& 
    \makecell[c]{$ -\left\|\mathbf{h}_m \circ \mathbf{r_1}_m ...\circ \mathbf{r_n}_m -\mathbf{t}_m\right\|_2- $\\$ \lambda\left\|\sin \left(\left(\mathbf{h}_p+\mathbf{r_1}_p...+\mathbf{r_n}_p-\mathbf{t}_p\right) / 2\right)\right\|_1 $}\\
    ComplEx & $\operatorname{Re}\left(h^{\top} \mathrm{diag}(r) \bar{t}\right)$ & $\operatorname{Re}\left(h^{\top} \mathrm{diag}(r_1) ...\mathrm{diag}(r_n) \bar{t}\right)$ \\
    DualE & $<Q_h \underline{\otimes} W_r^0, Q_t>$ & $<Q_h \underline{\otimes} W_{r_1}^0 ...\underline{\otimes} W_{r_n}^0, Q_t>$\\
    PairRE & $-\left\|\mathbf{h} \circ \mathbf{r}^H-\mathbf{t} \circ \mathbf{r}^T\right\|_1 $ & $-\left\|\mathbf{h} \circ \mathbf{r_1}^H ...\circ \mathbf{r_n}^H-\mathbf{t} \circ \mathbf{r_1}^T ... \circ \mathbf{r_n}^T\right\|_1 $ \\
    DistMult & $ \mathbf{h}^{\top}  \mathrm{diag}(\mathbf{r}) \mathbf{t} $ & $\mathbf{h}^{\top}  \mathrm{diag}(\mathbf{r_1}) ...\mathrm{diag}(\mathbf{r_n}) \mathbf{t} $\\
  \bottomrule
  \end{tabular}}}
\end{table}

We predict any missing fact jointly with the KGE score (Table~\ref{tab:score function}) and SPA score (Table~\ref{SPAscore}). For a query $(h,r,?)$, we substitute the tail entity with all candidate entities $(h,r,t')$ and compute their KGE scores $f(h,r,t)$ and compute the SPA score $s_p(h,r,t)$ if relation $r$ belongs to specific $p$ patterns. There may be many relations with the relation $r$ showing specific patterns, so we need to consider all of them. The final scores can be presented as:
\begin{equation}
\begin{split}
s(h,r,&t)=s_{kge}\left(h, r, t\right)+\\
&\lambda_{p} \frac{1}{\sum_{{\tau \in Set_{p}(r)}} MC_{\tau}} \sum_{{\tau \in Set_{p}(r)}} MC_{\tau}\left(s_p\left(h, r, t\right)-s_{kge}\left(h, r, t\right)\right) 
\end{split}
\end{equation}
where the function $s_{kge}$ represents the score function of KGE models and $s_p(h, r, t)$ represents the SPA score of the $p$ pattern. $Set_p(r)$ represents the $p$ pattern rule set of relation $r$. $\lambda_{p}$ represents the hyper-parameters of the $p$ pattern, and $MC_{\tau}$ represents the mean confidence of the rule $\tau$ to measure the credibility of the modified score.

\section{Experiments}\label{section5}
In this section, we introduce the experiment results with analysis. Specifically, in subsection~\ref{Evaluation Setup}, we introduce the datasets, metrics, and implementation first. Next in subsection~\ref{Quantitative Analysis over Relational Patterns}, we analyze the performance of the KGE models over relational patterns with three questions. Last, in subsection~\ref{Score-based Patterns Adaptation Results} and~\ref{Case Study}, we compare the performance of using SPA to optimize the score function on specific patterns and analyze SPA with the case study.

\subsection{Evaluation Setup}\label{Evaluation Setup}

\paragraph{Datasets.}
We have chosen two benchmark datasets, FB15k-237 and WN18RR, which are commonly used for evaluating Knowledge Graph Embedding (KGE) models. 
A summary of these datasets is presented in Appendix~\ref{dataset}.



\paragraph{Evaluation protocol.}\label{sec:protocol}
Following the same protocol as in TransE~\cite{bordes2013translating}, we evaluate link prediction performance by assessing the ranking quality of each test triple. For a given triple $(h,r,t)$ in the test set, we replace either the head $(h',r,t)$ or the tail entity $(h,r,t')$ with all entities and rank the candidate triples in descending order according to their scores. We select Mean Reciprocal Rank (MRR) as the evaluation metric. Furthermore, we use the ``filtered" setting~\cite{bordes2013translating} to eliminate the reconstructed triples that are already present in the KG.

 \paragraph{Implementation Details.}
In our experiments, we employ the combination of PCA confidence and HC thresholds to mine rules with AMIE3~\cite{lajus2020fast} and classify relations and triples then conduct quantitative analysis.
The best models are chosen by early stopping on the validation set using MRR over different patterns. As our SPA method optimizes the score function based on the embedding information from the trained KGE model without retraining or parameter adjustment, we do not modify the learning rate, margin, or other hyper-parameters. Instead, we only adjust the hyper-parameter $\lambda$ for the four patterns (\texttt{symmetric}, \texttt{inverse}, \texttt{multiple}, and \texttt{compositional}). Our settings for hyper-parameter selection are as follows: the hyper-parameters of \texttt{symmetric} pattern $\lambda_{sym}$, \texttt{inverse} pattern $\lambda_{inv}$, and \texttt{multiple} pattern $\lambda_{sub}$ are adjusted within the set $\{ \pm1,\pm2,\pm3,\pm4,\pm5,\pm10,\pm50,\pm100 \}$, while the hyper-parameter for the \texttt{compositional} pattern $\lambda_{comp2}$ is adjusted within the set $\{ \pm1\mathrm{e}{-5}, \pm1\mathrm{e}{-4}, \pm1\mathrm{e}{-3} $ \par \noindent$,\pm0.01, \pm0.02, \pm0.05, \pm0.1, \pm0.2, \pm0.5, \pm1, \pm2, \pm5 \}$. It should be noted that to ensure a fair comparison, experiments with the SPA strategy have the same hyper-parameters and implementation as the original models, except for pattern hyper-parameters. For more details about hyper-parameters in different KGE models, please refer to Appendix~\ref{sec:Hyper-parameter Settings}.

\begin{table}[h]
\centering
\caption{Statistics of the number of rules and relations for different relational patterns. \#Rule, \#r($P$) denote the number of the mined rules and relations in pattern $P$ $\#\{r|P \in Type(r)\}$ respectively. Sym, inv, mul, comp2, and comp3 are abbreviations for \texttt{symmetric}, \texttt{inverse}, \texttt{multiple}, and \texttt{compositional} patterns respectively. The mining results of relational patterns on FB15k-237 and WN18RR with five prescribed thresholds, where the symbol means the symbolization of different thresholds. i.e., the symbol of ``PCA=0.9 and HC=0.6 thresholds" can be written as $\theta_1$.
}
\setlength{\tabcolsep}{1mm}{ 
\label{tab:rules}
\resizebox{\columnwidth}{!}{
\begin{tabular}{@{}ccccccccc@{}}
\toprule
\multicolumn{3}{c|}{Confidence Thresholds} & \multicolumn{6}{c}{FB15k-237 / WN18RR} \\ \midrule
\multicolumn{1}{c|}{Symbol} & PCA & \multicolumn{1}{c|}{HC} & \# Rule & \#r(sym) & \#r(inv) & \#r(sub) & \#r(comp2) & \#r(comp3) \\ \midrule
\multicolumn{1}{c|}{$\theta_1$} & 0.9 & \multicolumn{1}{c|}{0.5} & 6k/5 & 6/3 & 4/0 & 14/0 & 49/0 & 111/1 \\
\multicolumn{1}{c|}{$\theta_2$} & 0.8 & \multicolumn{1}{c|}{0.5} & 6k/10 & 26/3 & 11/0 & 22/0 & 62/0 & 147/3 \\
\multicolumn{1}{c|}{$\theta_3$} & 0.6 & \multicolumn{1}{c|}{0.3} & 13k/41 & 28/3 & 26/0 & 32/0 & 96/0 & 203/7 \\
\multicolumn{1}{c|}{$\theta_4$} & 0.4 & \multicolumn{1}{c|}{0.1} & 39k/84 & 31/3 & 48/0 & 55/0 & 162/1 & 233/10 \\
\multicolumn{1}{c|}{$\theta_5$} & 0.2 & \multicolumn{1}{c|}{0.1} & 54k/115 & 31/3 & 67/0 & 73/0 & 185/1 & 233/10 \\ \bottomrule
\end{tabular}}
}
\end{table}

\subsection{Quantitative Analysis over Relational Patterns} \label{Quantitative Analysis over Relational Patterns}

Table~\ref{tab:rules} displays the number of rules mined in the two benchmarks at different confidence thresholds. 
We can obviously observe that the number of rules in FB15k-237 is much more than that in WN18RR, and there are all five relational patterns. While there are only two relational patterns in WN18RR, indicating that the data in FB15k-237 is more miscellaneous than that in WN18RR.


\begin{figure}
\centering
\includegraphics[width=5.5cm]{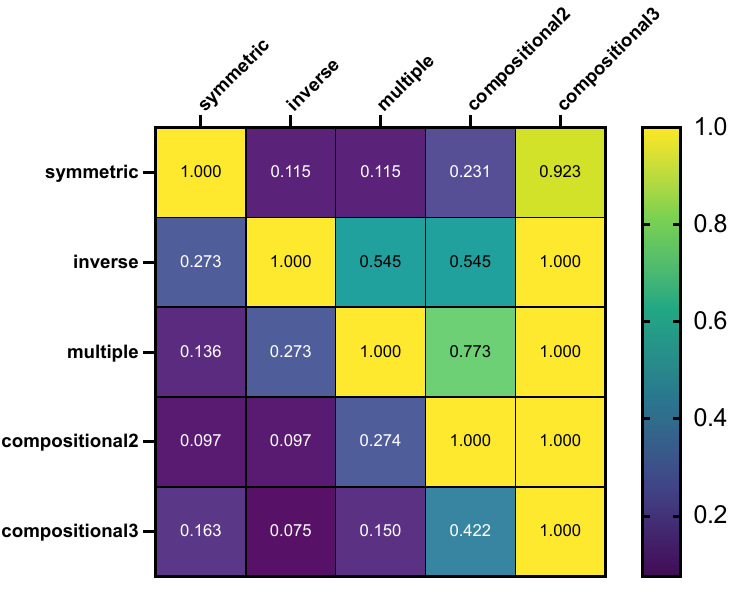}
\includegraphics[width=5.5cm]{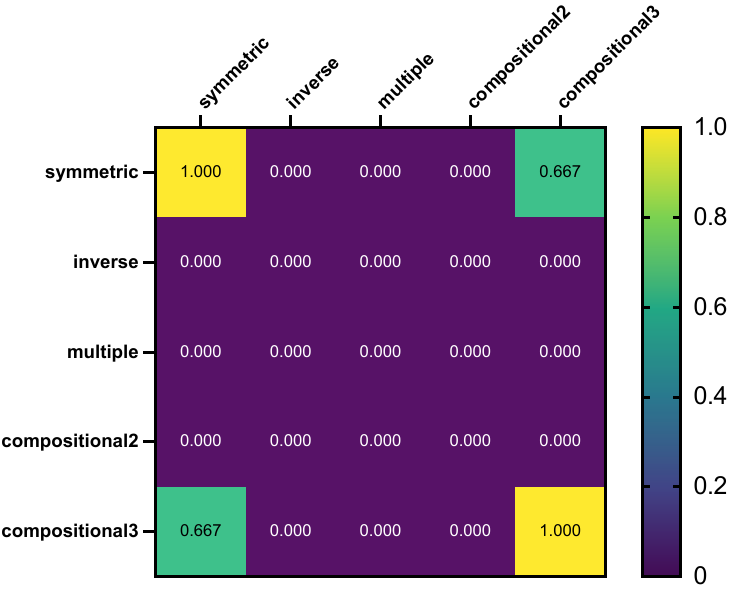}
\caption{The heat map of the relations' distribution in five relational patterns. 
From left to right are the statistical results under FB15k-237 and WN18RR datasets.
} \label{heat_map}
\end{figure}

Following the suggestions of RUGE~\cite{guo2018knowledge} and taking HC in the middle position, the evaluation of performance over patterns will be based on $\theta_2$ (PCA=0.8 and HC=0.5).

Fig.~\ref{heat_map} illustrates the heat map of relation distribution to show the overlapping between patterns. More details about the heat map can be found in Appendix~\ref{sec:Pattern Matrix}. In FB15k-237, 
the overlap between \texttt{compositional3} and others are close to 1, while the overlap between other relations is small, which means that relations in \texttt{compositional3} are massive.
The relationships in WN18RR are simpler without \texttt{inverse} or \texttt{multiple}. Furthermore, we closely examine the rules of \texttt{compositional3} in WN18RR and find that the body of rules consists of the head relation with two symmetric relations ~\cite{cui2022instance}, which causes the \texttt{compositional3} pattern to lose its compositional meaning.

Considering the significant overlap of \texttt{compositional3} with other patterns in FB15k-237 and its meaninglessness in WN18RR, we discard the \texttt{compositional3} pattern and conduct further experiments with the remaining four patterns.

Our analysis is driven by the form of three questions.


\textbf{Q1: Does a KGE model supporting a specific relational pattern in theory achieve better link prediction results on triples related to the relational pattern compared to another KGE model that does not support such a relational pattern?}

\begin{table}[]
\centering
\caption{Comparison of seven KGE models over various relational patterns with $\theta_2$. \ding{51} indicates that the model supports the pattern while \ding{55} indicates unsupported. 
The numbers with Star($*$) indicate the baseline in each pattern.
The percentages in parentheses represent the MRR ratios compared to the baseline while the radio of baseline in each pattern is always 100\%.
If the model gets better performance than the baseline, it indicates in \textbf{bold}, otherwise in \color[HTML]{777777}{gray}.
}
\renewcommand\arraystretch{1.1} 
\setlength{\tabcolsep}{1.5mm}{ 

\label{tab:Comparison}
\resizebox{.8\columnwidth}{!}{
\begin{tabular}{@{}clcccc@{}}
\toprule
\multicolumn{2}{c}{}                                 & \multicolumn{4}{c}{\textbf{Relation   Patterns}}                                       \\ \cmidrule(l){3-6} 
\multicolumn{2}{c}{\multirow{-2}{*}{\textbf{Model}}} & \texttt{\textbf{Symmetric}} & \texttt{\textbf{Inverse}} & \texttt{\textbf{Multiple}} & \texttt{\textbf{Composition2}} \\ \midrule
\multicolumn{2}{c}{\textbf{TransE}} &
  \textbf{\ding{55}(100\%)$*$} &
  {\color[HTML]{777777} \textbf{\ding{51}(98\%)}} &
  {\color[HTML]{1E1E1E} \textbf{\ding{55}(105\%)}} &
  {\color[HTML]{1E1E1E} \textbf{\ding{51}(119\%)}} \\
\multicolumn{2}{c}{\textbf{RotatE}} &
  {\color[HTML]{777777} \textbf{\ding{51}(98\%)}} &
  {\color[HTML]{777777} \textbf{\ding{51}(98\%)}} &
  {\color[HTML]{1E1E1E} \textbf{\ding{55}(102\%)}} &
  {\color[HTML]{1E1E1E} \textbf{\ding{51}(120\%)}} \\
\multicolumn{2}{c}{\textbf{HAKE}} &
  {\color[HTML]{777777} \textbf{\ding{51}(98\%)}} &
  {\color[HTML]{777777} \textbf{\ding{51}(96\%)}} &
  {\color[HTML]{1E1E1E} \textbf{\ding{55}(104\%)}} &
  {\color[HTML]{1E1E1E} \textbf{\ding{51}(124\%)}} \\
\multicolumn{2}{c}{\textbf{DistMult}} &
  {\color[HTML]{777777} \textbf{\ding{51}(94\%)}} &
  \textbf{\ding{55}(100\%)$*$} &
  {\color[HTML]{1E1E1E} \textbf{\ding{55}(106\%)}} &
  \textbf{\ding{55}(100\%)$*$} \\ 
\multicolumn{2}{c}{\textbf{ComplEx}} &
  {\color[HTML]{777777} \textbf{\ding{51} (61\%)}} &
  {\color[HTML]{777777} \textbf{\ding{51}(49\%)}} &
  {\color[HTML]{777777} \textbf{\ding{55}(52\%)}} &
  {\color[HTML]{1E1E1E} \textbf{\ding{55}(100\%)}} \\
\multicolumn{2}{c}{\textbf{DualE}} &
  {\color[HTML]{777777} \textbf{\ding{51}(90\%)}} &
  {\color[HTML]{777777} \textbf{\ding{51}(93\%)}} &
  \textbf{\ding{51}(100\%)$*$} &
  {\color[HTML]{1E1E1E} \textbf{\ding{51}(117\%)}} \\
\multicolumn{2}{c}{\textbf{PairRE}} &
  {\color[HTML]{1E1E1E} \textbf{\ding{51}(105\%)}} &
  {\color[HTML]{777777} \textbf{\ding{51}(99\%)}} &
  {\color[HTML]{1E1E1E} \textbf{\ding{51}(104\%)}} &
  {\color[HTML]{1E1E1E} \textbf{\ding{51}(125\%)}} \\\bottomrule
\end{tabular}}
}
\end{table}

We consider the answer to be \textbf{NO}. Intuition suggests that if a model itself does not support one relational pattern, it may have a lower MRR in the link prediction task compared to other KGE models that support the related relational pattern. However, the experimental results contradict our intuition.


Table~\ref{tab:Comparison} shows the comparison between seven KGE models. Due to the absence of some relational patterns in WN18RR, we only consider FB15k-237 for this experiment. 
In \texttt{symmetric}, the benchmark is TransE which does not support this pattern, while the other six KGE models (except PairRE) that do support it perform worse than TransE. The same result also appears in \texttt{inverse} and \texttt{multiple}. Only in \texttt{compositional2}, KGE models show significantly better performance for the supported than unsupported ones.
Moreover, it is difficult to judge the capability of KGE models over patterns when comparing those that support the same pattern. 
For instance, in \texttt{symmetric}, all supporting models perform around 90\% except for ComplEx, which also achieves 90\% in \texttt{inverse}, 105\% in \texttt{multiple}, and 120\% in \texttt{compositional2}. 


We consider that one of the reasons for this phenomenon is the complexity of the relations. The complex relatedness between relations makes relational patterns less discernible. As shown in Fig.~\ref{heat_map}, expect \texttt{compositional3}, the overlap between patterns is still not small and part overlaps values up to 50-70\%. This suggests that a single pattern is insufficient to construct the semantics of relations in complex KGs. Also,
the issue of sparsity presents a significant challenge for KGE models in reasoning processes~\cite{sharma2018towards,xiong2017explicit}. Generally, the infrequency of an entity correlates with diminished reasoning accuracy~\cite{zhang2019iteratively}. This prompts an investigation into the extent to which entity frequency impacts relational patterns, thus leading to our second research question.





\textbf{Q2: What is the impact of entity frequency for different KGE models over patterns?}

\begin{figure}[]
\centering
\includegraphics[height=4cm]{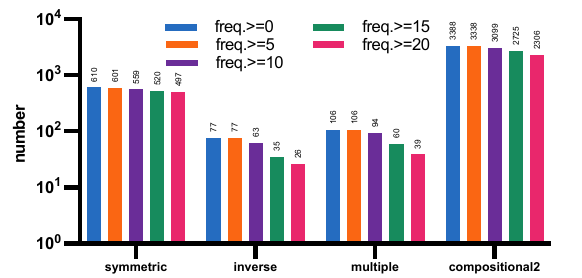}
\includegraphics[height=4cm]{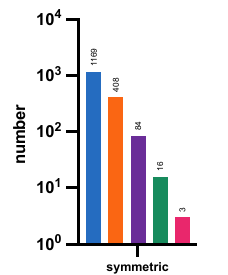}
    \caption{Statistics of triples in four patterns with entity frequency under $\theta_2$. From left to right are the results on FB15k-237 and WN18RR. (Note that the vertical axis is exponential)}\label{distribution}
\end{figure}


The statistics for the number of triples over patterns are shown in Fig.~\ref{distribution}. 
For a triple $(h, r, t)$, the threshold limit is the total number of occurrences of $h$ and $t$ in the KG. 
The \texttt{symmetric} pattern is significantly reduced from 0 to 5 frequency and the number of triples in the \texttt{compositional2} pattern is large with little change in entity frequency. 
In WN18RR, it only contains the \texttt{symmetric} pattern without \texttt{inverse} and \texttt{multiple} patterns. The number of \texttt{symmetric} triples is large and decreases sharply with increasing frequency. 
The subsequent experiments in this question will be based on the data mentioned above.

\begin{figure}[!bt]
\includegraphics[width=5.5cm]{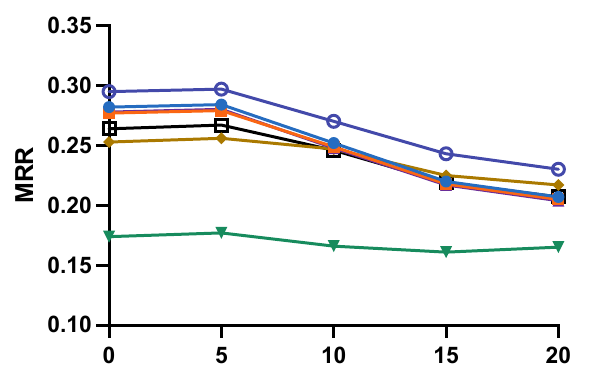}
\includegraphics[width=6.5cm]{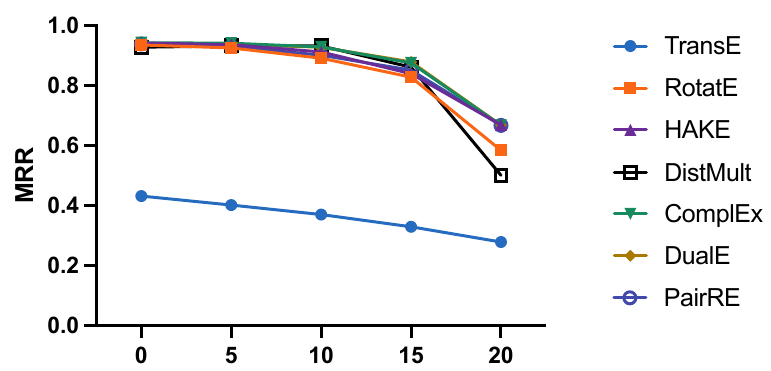}
\includegraphics[width=4cm]{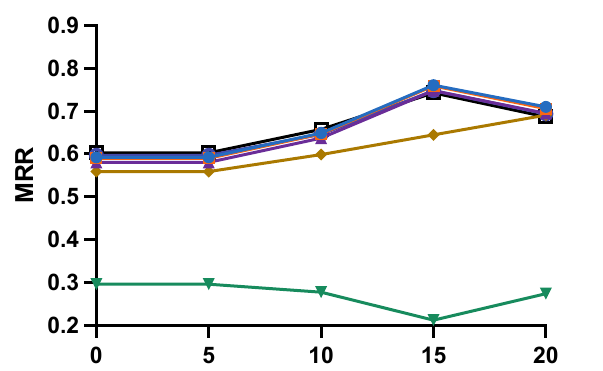}
\includegraphics[width=4cm]{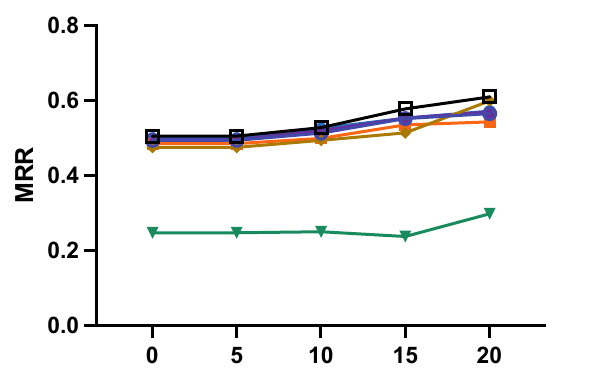}
\includegraphics[width=4cm]{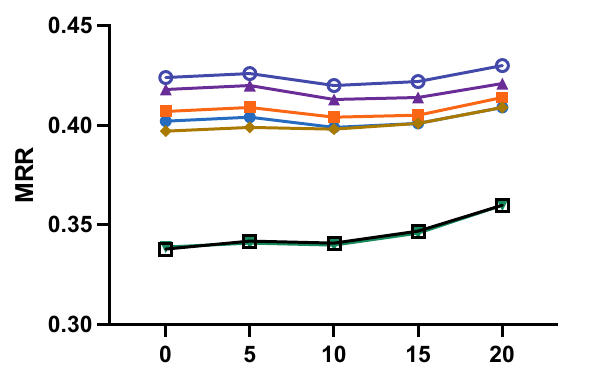}
    \caption{MRR of seven KGE models(TransE, RotatE, HAKE, DistMult, ComplEx, DualE, and PairRE) with varying entity frequency over patterns. The two figures in the first row, from left to right, are the experimental results of the \texttt{symmetric} pattern in FB15k-237 and WN18RR respectively. The second line from left to right is the results over \texttt{inverse}, \texttt{multiple}, and \texttt{compositional2} patterns in FB15k-237 respectively.}\label{frequency}
\end{figure}

Fig.~\ref{frequency} illustrates the performance of seven KGE models over different patterns with varying entity frequencies on two datasets. 
In FB15k-237, we observe that the performance of ComplEx is significantly worse than the other models. In WN18RR, TransE, which does not support \texttt{symmetric}, is considerably worse than the other models. Intuitively, we expect that as the constraints on entity frequency increase, the KGE model should be better at learning embeddings of entities and result in better performance. However, in the \texttt{symmetric} pattern, there is a noticeable downward trend as the constraints on entity frequency increase. The performance on \texttt{inverse} (the first plot in the second row of Fig.~\ref{frequency}) and \texttt{multiple} (the middle plot in the second row of Fig.~\ref{frequency}) in FB15k-237 is consistent with our intuition. Performance on \texttt{compositional2} shows an initial drop followed by a rise.

We find that the impact of entity frequency on different models (except ComplEx) is similar. Additionally, different patterns (except \texttt{symmetric}) generally improve as the constraint on entity frequency increases. We consider the reason for the downward trend as the increasing entity frequency is that the \texttt{symmetric} pattern relation solution is usually unitary. For example, in RotatE~\cite{sun2019rotate} and HAKE~\cite{zhang2020learning}, the symmetric relation tends to be $\pi $ to realize the rotation in complex space. With the increase in the entity frequency, the relations involved in the entities may also increase, resulting in the entity is not in the center symmetric position, which makes the model less effective.

Subsequently, we aim to examine the correlation between pattern-specific performance and overall performance in KGE inference, thereby formulating our final research question.

\textbf{Q3: If one KGE model performs better than another one on link prediction, is it because it improves uniformly across different relational patterns?}

\begin{figure}[]
\centering
\includegraphics[width=0.9\columnwidth]{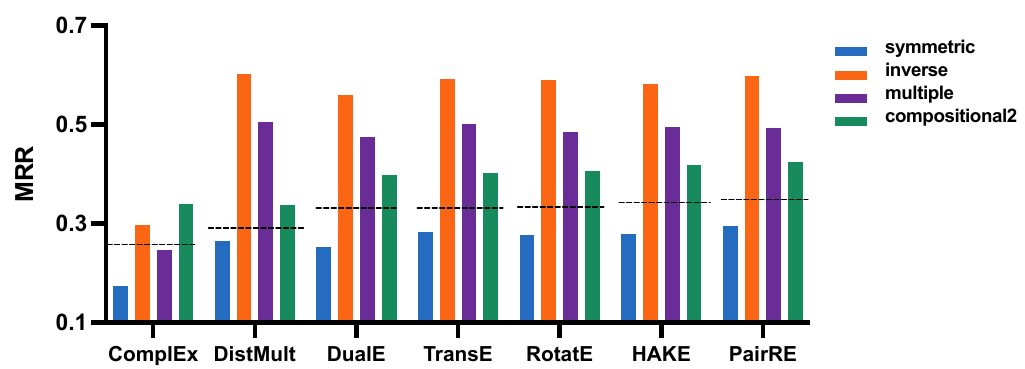}
    \caption{Comparison of seven KGE models under relational patterns with the overall performance in FB15k-237. The dotted lines in each model represent the performance under the complete test set of FB15k-237.}\label{comparison}
\end{figure}

We compare seven KGE models over relational patterns with the full set of MRR in FB15k-237, as shown in Fig.~\ref{comparison}. We find there are two scenarios. First, if one model demonstrates overall improvement compared to another, there will be significant enhancements in all relational patterns. For instance, when comparing ComplEx to DistMult, MRR has increased from 0.26 to 0.29, and it is evident that DistMult has improved in all patterns relative to ComplEx. Second, the relational pattern capability may be poor even when the overall performance is similar. Taking TransE, RotatE, and HAKE as examples, the overall performance of the three models is similar (0.33 for TransE, 0.333 for RotatE, and 0.329 for HAKE), but the capability of HAKE is slightly weaker than the other two models overall patterns.

\subsection{Score-based Patterns Adaptation Results} \label{Score-based Patterns Adaptation Results}
We compare the effectiveness of using SPA to optimize the score function on specific pattern datasets. We observe that different KGE models with SPA achieve greater performance in various pattern datasets on FB15K237 and WN18RR. Table~\ref{result_FB} displays the experimental result on FB15k-237.

\begin{table}[]
\centering
\caption{Link prediction results of several models evaluated on FB15k-237 with the metric of MRR. w/o SPA represents using only KGE score without SPA strategy and w SPA represents using SPA with KGE score. \textbf{Bold} numbers indicate the best performance over patterns.}
\label{result_FB}
\resizebox{\columnwidth}{!}{
\begin{tabular}{@{}ccccccccc@{}}
\toprule
\multirow{3}{*}{Model} 

& \multicolumn{8}{c}{FB15k-237} \\ \cmidrule(l){2-9} 

 & \multicolumn{2}{c|}{Sym. dataset} & \multicolumn{2}{c|}{Inv. dataset} & \multicolumn{2}{c|}{Sub. dataset} & \multicolumn{2}{c}{Comp.2 dataset} \\
 & w/o. SPA & \multicolumn{1}{c|}{w. SPA} & w/o. SPA & \multicolumn{1}{c|}{w. SPA} & w/o. SPA & \multicolumn{1}{c|}{w. SPA} & w/o. SPA & w. SPA \\ \midrule
\multicolumn{1}{c|}{TransE} & .282 & \multicolumn{1}{c|}{\textbf{.301}} & .592 & \multicolumn{1}{c|}{\textbf{.649}} & .500 & \multicolumn{1}{c|}{\textbf{.546}} & .402 & \textbf{.403} \\
\multicolumn{1}{c|}{RotatE} & .277 & \multicolumn{1}{c|}{\textbf{.293}} & .590 & \multicolumn{1}{c|}{\textbf{.644}} & .485 & \multicolumn{1}{c|}{\textbf{.543}} & \textbf{.407} & \textbf{.407} \\
\multicolumn{1}{c|}{HAKE} & .278 & \multicolumn{1}{c|}{\textbf{.288}} & .581 & \multicolumn{1}{c|}{\textbf{.633}} & .495 & \multicolumn{1}{c|}{\textbf{.559}} & \textbf{.418} & \textbf{.418} \\
\multicolumn{1}{c|}{DistMult} & \textbf{.264} & \multicolumn{1}{c|}{\textbf{.264}} & .603 & \multicolumn{1}{c|}{\textbf{.622}} & .505 & \multicolumn{1}{c|}{ \textbf{.529}} & \textbf{.338} & \textbf{.338} \\ 
\multicolumn{1}{c|}{ComplEx} & .174 & \multicolumn{1}{c|}{\textbf{.179}} & \textbf{.296} & \multicolumn{1}{c|}{\textbf{.296}} & .247 & \multicolumn{1}{c|}{\textbf{.273}} & \textbf{.339} & \textbf{.339} \\
\multicolumn{1}{c|}{DualE} & .253 & \multicolumn{1}{c|}{\textbf{.254}} & .559 & \multicolumn{1}{c|}{\textbf{.564}} & .475 & \multicolumn{1}{c|}{\textbf{.483}} & .397 & \textbf{.399} \\
\multicolumn{1}{c|}{PairRE} & .295 & \multicolumn{1}{c|}{\textbf{.316}} & .597 & \multicolumn{1}{c|}{\textbf{.676}} & .494 & \multicolumn{1}{c|}{\textbf{.555}} & .424 & \textbf{.425} \\\bottomrule
\end{tabular}
}
\end{table}

On FB15k-237, our SPA strategy attains the best performance, demonstrating that it can effectively learn all patterns in specific pattern datasets. In the performance comparison of the \texttt{symmetric} pattern, KGE models with SPA, such as TransE, RotatE, and PairRE, achieve a 2\% improvement in MRR. For the \texttt{inverse} pattern, KGE models with SPA, such as TransE, RotatE, HAKE, and PairRE, show MRR improvements of 5-8\%. For the \texttt{multiple} pattern, KGE models with SPA, such as TransE, RotatE, HAKE, and PairRE, exhibit MRR improvements of around 6\%. For the \texttt{compositional2} pattern, we find that SPA gets only slight improvements in DualE and PairRE. A possible explanation for this phenomenon is that during the experiment, we calculate the contribution of the \texttt{compositional2} pattern in the form of 2-hops, which may further amplify errors and make the effect less pronounced. The experimental result of WN18RR is presented in Appendix~\ref{sec:SPA Results on WN18RR} and is consistent with our conclusions in FB15k-237.

\subsection{Case Study} \label{Case Study}
Score-based Patterns Adaptation (SPA) utilizes the information from relational patterns, which is simple but effective to be applied to KGE models without further training. As Table~\ref{case} shows, we provide an intuitive demonstration for SPA. We select an example case (1980 Summer Olympics, /olympics/olympic\_games/ sports, ?) from the FB15k-237 and list the tail prediction rank on $s_{kge}(h,r,t)$, $s_{p}(h,r,t)$, and $s(h,r,t)$ based on RotatE.

\begin{table}[]
\centering
\caption{The rank of part entities in different score functions. The reasoning task is 
(1980 Summer Olympics, /olympics/olympic\_games/sports, ?) from the FB15k-237,  and we list the tail prediction rank in KGE score $s_{kge}(h,r,t)$, SPA score $s_{p}(h,r,t)$, and KGE with SPA score $s(h,r,t)$ based on RotatE. 
The rankings are sorted from highest to lowest with ``filter" and the ``\textbf{Rowing}" entity is the grounding.}
\renewcommand\arraystretch{1.2} 
\resizebox{0.95\columnwidth}{!}{
\label{case}
\setlength{\tabcolsep}{1.7mm}{ 
\begin{tabular}{@{}cccc@{}}
\toprule
\multicolumn{4}{c}{Q: (1980 Summer Olympics, /olympics/olympic\_games/sports, ?) A:   \textbf{Rowing}} \\\midrule
Rank   & KGE score $s_{kge}(h,r,t)$     & SPA score $s_{p}(h,r,t)$    & KGE and SPA $s(h,r,t)$    \\ \midrule
1 & {\color[HTML]{9C9C9C} \textbf{Artistic gymnastics}} & {\color[HTML]{9C9C9C} \textbf{Artistic gymnastics}} & Canoe Slalom                                        \\
2 & {\color[HTML]{828282} \textbf{Swimming}}            & {\color[HTML]{828282} \textbf{Swimming}}            & {\textbf{Rowing}}              \\
3 & {\color[HTML]{696969} \textbf{Volleyball}}          & Freestyle wrestling                                 & {\color[HTML]{828282} \textbf{Swimming}}            \\
4 & Freestyle wrestling                                 & {\color[HTML]{696969} \textbf{Volleyball}}          & {\color[HTML]{9C9C9C} \textbf{Artistic gymnastics}} \\
{\color[HTML]{000000} 5} & { \textbf{Rowing}} & Archery                                & Cycling                                    \\
6                        & Archery                                & { \textbf{Rowing}} & {\color[HTML]{696969} \textbf{Volleyball}} \\ \bottomrule
\end{tabular}}}
\end{table}

Firstly, we get the rule $\tau$: /user/jg/default\_domain/olympic\_games/sports($H$, $T$) $\rightarrow$ /olympics/olympic\_games/sports$(H,T)$ with PCA $=$ $0.83$ and HC $= 0.72$ from test set which means relations in the head and body of rule $\tau$ with high similarity. The rank of the correct answer (Rowing) is 5,6, and 2 with different scores respectively, while confusing answers such as Artistic gymnastics, Swimming, and Volleyball are decreased in the rank of the $s(h,r,t)$, which means that SPA calculates the SPA score from the body of rules with the KGE score can achieve more accurate in reasoning.

%
%
%

\section{Conclusion and Future Work}\label{section6}
We study KGC tasks in KGE models based on relational patterns and analyze the result in theory, entity frequency, and part-to-whole three aspects with some counterintuitive and interesting conclusions. 
Firstly, theoretical backing for a KGE model's relational pattern doesn't ensure its superiority over models without such support. Secondly, entity frequency differently affects relational patterns' performance; it decreases for symmetric patterns yet increases for others with rising frequency.
Finally, a significantly outperforming KGE model consistently excels across all relational patterns otherwise doesn't.

In the future, we believe that the research of KGE should not be confined to relational patterns.
Greater attention should be given to analyzing the correlations between overall relations from a macro perspective, negative sampling~\cite{kamigaito2022comprehensive,hajimoradlou2022stay}, and loss function.

\paragraph*{Supplemental Material Statement:}
Our source code, datasets and results of study with SPA are all available from GitHub at \url{https://github.com/zjukg/Comprehensive-Study-over-Relational-Patterns}.

\subsubsection*{Acknowledgements}
This work is funded by Zhejiang Provincial Natural Science Foundation of China (No. LQ23F020017), Yongjiang Talent Introduction Programme (2022A-238-G), the National Natural Science Foundation of China (NSFCU19B2027, NSFC91846204), joint project DH-2022ZY0012 from Donghai Lab.

\bibliographystyle{splncs04}
\bibliography{references}

\newpage
\section*{Appendix}
\appendix

\section{Implementation Details on Compositional Pattern.} \label{sec:Compositional Pattern}
The \texttt{compositional} pattern differs from other patterns in that its rule length is no longer 1. We mainly discuss rules with rule length 2. Since the rules in the \texttt{compositional} pattern are usually not close-path rules, we need to get their inverse relation and define the function of compositional inference. Our approach is mainly divided into the following two steps: 1)construction of inverse relation for all relations. 2) Define the function of the compositional inference.

\paragraph{Construction of inverse relation for all relations} To make the experiment simpler, we do not add a new reverse triple $(t, r^{-1}, h)$ of $(h,r,t)$ in the dataset, but derive its inverse relation from the properties of itself. Table~\ref{Reverse} lists all the inverse relations that need to be used in our experiment.

\begin{table}[]
\centering
\caption{Reverse Relation for different KGE models. Note that TransE, RotatE, DualE represent entities and relations into real, complex, and dual-quaternion space respectively, inverse relations of them the inverse in the corresponding space. HAKE is the combine of TransE and RotatE. The relation in DistMult and ComplEx are diagonal matrixes and the inverse of relations are the inverse of the matrixes.}
\label{Reverse}
\renewcommand\arraystretch{1.2} 
\begin{tabular}{@{}cc@{}}
\toprule
Model    & Reverse Relation($r^{-1}$)                          \\ \midrule
TransE   & $\textbf{r}^{-1}=-\textbf{r}$                                         \\
RotatE   & $\textbf{r}^{-1}=\overline{\textbf{r}}$                               \\
HAKE     & $\textbf{r}_{m}^{-1}=-\textbf{r}_{m} \quad \textbf{r}_{p}^{-1}=\overline{\textbf{r}}$   \\
DistMult & $\mathrm{diag}(r)^{-1}=1/\mathrm{diag}(r)$                            \\
ComplEx  & $\mathrm{diag}(r)^{-1}=1/\mathrm{diag}(r)$                            \\
DualE    & $W_{r}^{\Diamond   -1}=\overline{W_{r}^{\Diamond}}$ \\
PairRE   & $\textbf{r}^{H-1}=1/r^{H} \quad r^{T-1}=1/\textbf{r}^{T}$             \\ \bottomrule
\end{tabular}
\end{table}

\paragraph{Define the function of the compositional inference}
The original compositional rule can be encoded into a closed-path rule by inverse relation. Table~\ref{tab:score function} defines the compositional inference function $Path(h,r_1,...r_n,t)$, the proof of them can be listed as follows:

For TransE, $r_1$,$r_2$ can compose $r$, we can get $\textbf{r}_1 + \textbf{r}_2 = \textbf{r}$.

\textit{Proof:} If $r_1(h,e_1)$, $r_2(e_1,t$), and $r(h,t)$, we have 
\begin{equation}
\textbf{h} + \textbf{r}_1 = \textbf{e}_1 \wedge \textbf{e}_1 + \textbf{r}_2 = \textbf{t} \wedge \textbf{h} + \textbf{r} = \textbf{t} \Rightarrow \textbf{r}_1+\textbf{r}_2=\textbf{r}
\end{equation}

For RotatE, HAKE, DualE, and PairRE, the proof of the compositional inference function $Path(h,r_1,...r_n,t)$ of relationships between relation in head $r$ and relations in body $r_1,...r_n$ can be found in works~\cite{sun2019rotate,zhang2020learning,cao2021dual,chao2020pairre} respectively. For DistMult and ComplEx, 
since these two models themselves do not support \texttt{compositional} pattern, we use the method of modeling as a multiplication of matrices proposed by Yang~\cite{yang2014embedding} to construct compositional inference function $Path(h,r_1,...r_n,t)$.

\section{Dataset Detail} \label{dataset}

Table~\ref{tab:dataset} shows the detail of FB15k-237 and WN18RR.

\begin{table}[h]
  \centering
  \setlength{\tabcolsep}{2mm}{ 
  \centering
  \caption{Statistics of datasets. The symbols \#E and \#R denote the number of entities and relations respectively. \#TR, \#VA, and \#TE denote the size of the train set, validation set, and test set respectively }
  \label{tab:dataset}
  \begin{tabular}{cccccc}
    \toprule
    Dataset & \#E & \#R & \#TR & \#VA & \#TE\\
    \midrule
    FB15k-237 & 14541 & 237 & 272115 & 17535 & 20466 \\
    WN8RR     & 40493 & 11  & 86835  & 3034  & 3134  \\
  \bottomrule
  \end{tabular}
  }
\end{table}

\section{The Hyper-parameter Settings.} \label{sec:Hyper-parameter Settings}
Table~\ref{parameters} shows hyper-parameters settings of KGE models over different pattern datasets.

\begin{table}[]
\centering
\caption{Hyper-parameters settings of KGE models over different pattern}
\label{parameters}
\renewcommand\arraystretch{1} 
\setlength{\tabcolsep}{1.5mm}{ 
\begin{tabular}{@{}c|cccc@{}}
\toprule
\multicolumn{1}{l|}{KGE model} & $\lambda_{sym}$ & $\lambda_{inv}$ & $\lambda_{mul}$ & $\lambda_{comp}$ \\ \midrule
TransE   & -2  & -2 & -3 & 0.2  \\
RotatE   & -4  & -1 & -4 & -0.01 \\
HAKE     & -2  & -1 & -3 & 0.1   \\
DistMult & -2  & -2 & -4 & 1e-5    \\
ComplEx  & -2  & -1 & -4 & -0.01      \\
DualE    & -2  & -1 & -3 & -0.01   \\
PairRE   & -10 & -2 & -2 & 0.5   \\ \bottomrule
\end{tabular}}
\end{table}

\section{Pattern Matrix} \label{sec:Pattern Matrix}
The relationships between relations may be highly complex, and overlapping between patterns can affect the final analysis, we propose the \textbf{Pattern Matrix} to indicate the degree of intersection between pairs of relational patterns. 
Each element $M_{ij}$ of the matrix $M$ represents 
the ratio of the relation overlap of relational pattern $P_i$ to $P_j$ that
\begin{equation}
M_{ij} = \frac{| R_{P_i} \cap R_{P_j} |}{| R_{P_i} |}
,  \;\;\; R_{P_k} = \{ r | P_k \in Type(r) \}
\end{equation}
where $R_{P_i}$ represents the set of relations that belong to $P_i$, $| X |$
represents the length of the set $X$, and $\cap$ denotes the intersection of sets.

\section{SPA Results on WN18RR.} \label{sec:SPA Results on WN18RR}
Table~\ref{result_WN} displays the experimental results on WN18RR. On WN18RR, which only has a \texttt{symmetric} pattern due to the absence of complex connections between relations like in FB15k-237, we analyze the \texttt{symmetric} pattern. With SPA, TransE achieves a 40\% improvement in MRR, and by comparing the Hit@N metrics, the primary improvement is in Hit@1, which increases from 0 to 0.8 with SPA. This may be because TransE does not support the symmetric model, and there is an evident symmetric relationship in WN18RR, causing TransE to learn the symmetric relation as a \textbf{0} vector with high probability, resulting in low performance in Hit@1. Except for TransE, the less conspicuous performance of other models may also be related to the obvious symmetric relationship in the dataset itself, which enables different models that support the \texttt{symmetric} pattern to better learn symmetric relationships.


\begin{table}[]
\centering
\caption{Link prediction results of several models evaluated on the WN18RR dataset with the metric of MRR. 
\textbf{Bold} numbers indicate the best performance over patterns.}
\label{result_WN}
\begin{tabular}{@{}cccc@{}}
\toprule
\multirow{2}{*}{Model} & \multicolumn{2}{c}{Symmetry dataset} & \multicolumn{1}{l}{Hyper-parameters} \\ \cmidrule(l){2-4} 
                              & w/o. SPA & w. sym. SPA & $\lambda_{sym}$ \\ \midrule
\multicolumn{1}{c|}{TransE}   & .431    & \textbf{.833}       & 50            \\
\multicolumn{1}{c|}{RotatE}   & \textbf{.935}    & \textbf{.935}       & 1           \\
\multicolumn{1}{c|}{HAKE}     & \textbf{.942}    & \textbf{.942}       & 1          \\
\multicolumn{1}{c|}{DistMult} & \textbf{.928}    & \textbf{.928}       & 1           \\
\multicolumn{1}{c|}{ComplEx}  & .943    & \textbf{.944}       & 1             \\
\multicolumn{1}{c|}{DualE}    & .938    & \textbf{\textbf{.939}}       & 2             \\
\multicolumn{1}{c|}{PairRE}   & \textbf{.939}    & \textbf{.939}       & 1            \\ \bottomrule
\end{tabular}
\end{table}

\end{document}